\documentclass{article}

\usepackage{arxiv}

\newif\ifarxiv
\arxivtrue

\usepackage[utf8]{inputenc}
\usepackage[T1]{fontenc}
\usepackage{url}
\usepackage{booktabs}
\usepackage{microtype}
\usepackage{graphicx}
\usepackage{algorithm}
\usepackage{amsmath,amsthm}
\usepackage{newtxtext,newtxmath}
\usepackage{textcomp}
\usepackage{xcolor}
\usepackage{algpseudocode}
\usepackage{multicol}
\usepackage{multirow}
\usepackage{tabularx}
\usepackage{enumitem}
\usepackage{bm}
\usepackage[style=authoryear,natbib]{biblatex}
\usepackage[bookmarks=false,hypertexnames=true,hidelinks]{hyperref}

\DeclareSourcemap{  
  \maps[datatype=bibtex]{
    \map{
      \step[fieldsource=doi,final]
      \step[fieldset=url,null]
      \step[fieldset=eprint,null]
    }                   
  }
}

\graphicspath{{./figures/}}

\newtheorem{definition}{Definition}

\newcolumntype{R}{>{\raggedleft\arraybackslash}X}
\newcolumntype{L}{>{\raggedright\arraybackslash}X}

\title{Hard to Forget: Poisoning Attacks on Certified Machine Unlearning}

\date{February 10, 2022}

\author{
  Neil G.~Marchant \\
  School of Computing and Information Systems\\
  University of Melbourne\\
  \texttt{nmarchant@unimelb.edu.au}
  \And
  Benjamin I.~P.~Rubinstein \\
  School of Computing and Information Systems\\
  University of Melbourne\\
  \texttt{brubinstein@unimelb.edu.au}
  \AND
  Scott Alfeld\\
  Department of Computer Science\\
  Amherst College\\
  \texttt{salfeld@amherst.edu}
}

\renewcommand{\headeright}{}
\renewcommand{\undertitle}{}
\renewcommand{\shorttitle}{Poisoning Attacks on Certified Machine Unlearning}

\hypersetup{
  pdftitle={Hard to Forget: Poisoning Attacks on Certified Machine Unlearning},
  pdfauthor={Neil G. Marchant, Benjamin I. P. Rubinstein, Scott Alfeld},
  pdfkeywords={machine unlearning, data poisoning, adversarial machine learning},
}

\newif\ifincludeapp
\includeapptrue

\newif\ifsqueeze
\squeezetrue

\newcommand{\learn}{\ensuremath{A}}
\newcommand{\unlearn}{\ensuremath{M}}
\newcommand{\hessian}{\mathrm{H}}
\newcommand{\infl}{\mathcal{I}}
\newcommand{\jac}{\mathrm{J}}
\newcommand{\data}{\ensuremath{D}}
\newcommand{\pois}{\mathrm{psn}}
\newcommand{\clean}{\mathrm{cln}}

\newcommand{\prob}{\ensuremath{P}}
\newcommand{\euler}{e} %

\renewcommand{\vec}[1]{\bm{\mathrm{#1}}}
\newcommand{\mat}[1]{\mathbf{#1}}
 
\newcommand{\E}{\mathbb{E}}
\newcommand{\reals}{\ensuremath{\mathbb{R}}}

\newcommand{\Xspace}{\ensuremath{\mathcal{X}}}
\newcommand{\Yspace}{\ensuremath{\mathcal{Y}}}
\newcommand{\Zspace}{\ensuremath{\mathcal{Z}}}
\newcommand{\hyposet}{\ensuremath{\mathcal{H}}}
\newcommand{\dataspace}{\ensuremath{\Zspace^\ast}}
\newcommand{\risk}{R}

\newcommand{\secref}[1]{Sec.~\ref{#1}}
\newcommand{\alref}[1]{Alg.~\ref{#1}}

\begin{document}
\maketitle

\begin{abstract}
  The right to erasure requires removal of a user's information from data held 
by organizations, with rigorous interpretations extending to downstream 
products such as learned models. 
Retraining from scratch with the particular user's data omitted fully removes 
its influence on the resulting model, but comes with a high computational cost.
Machine ``unlearning'' mitigates the cost incurred by full retraining: instead, 
models are updated incrementally, possibly only requiring retraining when 
approximation errors accumulate. 
Rapid progress has been made towards privacy guarantees on the 
indistinguishability of unlearned and retrained models, but current formalisms 
do not place practical bounds on computation.
In this paper we demonstrate how an attacker can exploit this oversight, 
highlighting a novel attack surface introduced by machine unlearning.
We consider an attacker aiming to increase the computational cost of data 
removal. 
We derive and empirically investigate a poisoning attack on certified machine 
unlearning where strategically designed training data triggers complete 
retraining when removed.

\end{abstract}


\section{Introduction}

Much of modern machine learning has been advanced by the availability of large 
volumes of data. 
However, organizations that collect user data must comply with data protection 
regulations and often prioritize privacy to mitigate risk of civil litigation 
and reputational damage. 
Prominent regulations---i.e., GDPR \citep{eu2016gdpr} and CCPA 
\citep{ccpa2018}---encode a \emph{right to erasure}---organizations must 
``erase personal data without undue delay'' when requested. 
When tested, such regulations have been found to extend to products 
derived from personal data, including trained models~\citep{ftc2021}. 
\citet{veale2018algorithms} argue that models could be legally classified as 
personal data, which agrees with the draft guidance from the 
\citet{ico2020guidance} that ``erasure of the data... may not be possible 
without retraining the model... or deleting the model altogether''.
Such concerns are justified by a legitimate privacy risk: training data can 
be practically reconstructed from models \citep{fredrikson2015model, 
shokri2017membership, veale2018algorithms}.

Na\"ively erasing data from a model by full retraining can be computationally 
costly. 
This has motivated the recent study of \emph{machine unlearning}: how to 
efficiently undo the impact of data on a model. 
\citet{guo2020certified} develop an approximate update which is 3 orders of 
magnitude faster than retraining, or supporting up to 10,000 approximate 
updates before retraining is required, with only a small drop in accuracy. 
\citet{ginart2019making} find a $100\times$ improvement in deletion efficiency 
with comparable clustering quality. 
Theoretical investigations has led to guarantees on privacy 
(indistinguishability of unlearned models with retrained models), but have 
largely ignored computation cost.

In an adversarial setting where risks to privacy motivate unlearning, we 
advocate that computation should also be viewed under an adversarial lens. 
While the aforementioned empirical results paint an optimistic picture for 
machine unlearning, such work assumes passive users who are trusted: they 
work within provided APIs, but also don't attempt to harm system performance 
in any way. 
In this paper we consider an active adversary as is more typical in the 
adversarial learning field~\citep{huang2011adversarial}. 
Users in this work can issue strategically chosen data for learning, and 
then request their data be unlearned. We adopt a typical threat model where 
users may apply a bounded perturbation to their (natural) data. 
Boundedness is motivated by a desire to remain undetected. 
Under this threat model \textbf{we identify a new vulnerability: poisoning 
attacks that effectively slow down unlearning.}

We highlight several perspectives on our results. 
They call into question the extent to which unlearning improves performance 
over full retraining which achieves complete indistinguishability.
Our work highlights the risk of deploying approximate unlearning algorithms 
with data-dependent running times. 
And more broadly we show that data poisoning can harm \emph{computation} 
beyond only accuracy---analogous to conventional denial-of-service attacks.

\paragraph{Contributions.} 
We identify a new vulnerability of machine learning systems. 
Our suite of experiments considers: 
the attack's effect in a wide range of parameter settings; 
both white-box and grey-box attacks; 
a range of perturbation geometries and bounds; 
trade-offs between optimality of the attacker's objective and time to compute 
the attack; 
and the feasibility of running the attack long term.

\section{Preliminaries}
This section sets the scene for our attack. 
We introduce the target of our attack in 
\secref{sec:unlearning-problem}---learning systems that support data 
erasure requests via machine unlearning. 
We then review a certified machine unlearning method that we use to 
demonstrate our attack. 
\secref{sec:threat-model} closes with a discussion of the vulnerability 
and our assumed threat model. 
Table~\ref{tbl:notation} summarizes notation used throughout the paper.

\subsection{Machine learning on user data}
\label{sec:unlearning-problem}
Consider an organization that trains machine learning models using data 
from consenting users. 
In line with privacy regulations, the organization allows users to revoke 
consent at any time and submit an erasure request for some (or all) of their 
data.
When fulfilling an erasure request, the organization takes a cautious approach 
to user privacy---erasing the user's raw data and removing the data's 
influence on trained models.
We assume erasing raw data (e.g.\ from a database) is straightforward and 
focus here on the task of removing data from a trained model.

Formally, let $\dataspace$ be the space of training datasets consisting of 
samples from an example space $\Zspace$.
Given a training dataset $\data \in \dataspace$, the organization deploys 
a \emph{learning algorithm} that returns a model $\learn(D)$ in hypothesis space 
$\hyposet$.\footnote{
  In practice, a ``model'' may contain parameters used at inference time, 
  as well as metadata used to speed up unlearning.
  For notational brevity, we absorb the metadata into the definition of 
  $\hyposet$.
}
To support erasure requests, the organization also deploys an \emph{unlearning 
algorithm} $\unlearn$ which, given a training dataset $\data \in \dataspace$, a 
subset $\data' \subset \data$ to erase and a model $h \sim \learn(\data)$, 
returns a sanitized model $\unlearn(\data, \data', h) \in \hyposet$ that is 
approximately (to be made mathematically precise below) 
$\learn(\data \setminus \data')$.
We allow $\learn$ and $\unlearn$ to be randomized algorithms---each can be interpreted as 
specifying a distribution over $\hyposet$ conditioned on their inputs.

To ensure no trace of the erased data is left behind, we assume $\learn$ and $\unlearn$ 
satisfy a rigorous definition of unlearning.
Specifically, we assume the distribution of unlearned models 
$\unlearn(\data, \data', \learn(\data))$ is \emph{statistically indistinguishable} from 
the distribution of models retrained on the remaining data 
$\learn(\data \setminus \data')$. 
This is known as \emph{approximate unlearning} and there are variations 
depending on how statistical indistinguishability is defined.
For concreteness, we adopt the unlearning definition and algorithms of
\citet{guo2020certified}, which we review in the next section. 

\begin{table}
  \caption{Summary of notation.}
  \label{tbl:notation}
  \ifarxiv \else \footnotesize \fi 
  \begin{tabularx}{\linewidth}{lX}
    \toprule
    Notation & Explanation \\
    \midrule
    $\learn$ & 
      Learning algorithm \\
    $\unlearn$ & 
      Unlearning algorithm \\
    $\data$, $\data_\pois$, $\data_\clean$ & 
      Dataset (generic, poisoned, clean) \\
    $\vec{X}$, $\vec{X}_\pois$, $\vec{X}_\clean$ &
      Feature matrix (generic, poisoned, clean) \\
    $\hyposet$ &
      Hypothesis space for model \\
    $\vec{\theta}$ &
      Model parameters \\
    $R$, $R_{\vec{b}}$ & 
      Learning objective (unperturbed, perturbed) \\
    $\vec{b}$ & 
      Objective perturbation coefficients \\
    $\sigma$ & 
      Standard deviation of $\vec{b}$ \\
    $\lambda$ & 
      Regularization strength \\
    $\ell$ & 
      Loss function \\
    $\gamma$ & 
      Lipschitz constant of $\ell''$ \\
    $\epsilon, \delta$ & 
      Indistinguishability parameters \\
    $\beta$ & 
      Bound on gradient norm of $R_{\vec{b}}$ \\
    $C$ & 
      Attacker's estimate of unlearning cost \\
    $\{g_j\}$ & 
      Attacker's inequality constraints \\
    \bottomrule
  \end{tabularx}
\end{table}

\subsection{Certified removal}
\label{sec:cr}

This section reviews learning\slash unlearning algorithms for linear 
models proposed by \citet{guo2020certified}. 
The algorithms satisfy a definition of approximate unlearning called 
\emph{certified removal}, where indistinguishability is defined in a similar 
manner as $(\epsilon, \delta)$-differential privacy \citep{dwork2006our}.

\begin{definition}%
  Given $\epsilon, \delta > 0$ a removal mechanism $\unlearn$ performs 
  $(\epsilon, \delta)$-certified removal for learning algorithm $\learn$ if 
  $\forall \mathcal{T} \subseteq \hyposet, \data \in \dataspace, 
  \vec{z} \in \data$:
  \begin{equation*}
    \prob(\unlearn(\data, \{\vec{z}\}, \learn(\data)) \in \mathcal{T}) 
      \leq \euler^{\epsilon} \prob(\learn(\data \setminus \{\vec{z}\}) \in \mathcal{T}) + \delta 
  \end{equation*}
  and
  \begin{equation*}
    \prob(\learn(\data \setminus \{\vec{z}\}) \in \mathcal{T})
      \leq \euler^{\epsilon} \prob(\unlearn(\data, \{\vec{z}\}, \learn(\data)) \in \mathcal{T})  + \delta.
  \end{equation*}
\end{definition}

The learning\slash unlearning algorithms we consider support linear models that minimize 
a regularized empirical risk of the form:
\begin{equation}
  \risk(\vec{\theta}; \data) = \sum_{(\vec{x}, y) \in \data} \left\{ 
      \ell(\vec{\theta}^\top \vec{x}, y) + \frac{\lambda}{2} \| \vec{\theta} \|_2^2 
        \right\}
  \label{eqn:cr-risk}
\end{equation}
where $\vec{\theta} \in \reals^d$ is a vector of learned weights, 
$\ell: \reals \times \reals \to \reals$ is a convex loss that is differentiable 
everywhere, and $\lambda > 0$ is a regularization hyperparameter.
Note that we are assuming the instance space $\Zspace$ is the product of a 
feature space $\Xspace \subseteq \reals^d$ and a label space 
$\Yspace \subseteq \reals$. 
Thus supervised regression and classification are supported.

\paragraph{Learning algorithm.}
\label{sec:cr-learning}
Since unlearning is inexact (see below), random noise is injected during 
learning to ensure indistinguishability.
Concretely, the original minimization objective $\risk(\vec{\theta}; \data)$ is 
replaced by a perturbed objective
\begin{equation}
  \risk_{\vec{b}}(\vec{\theta}; \data) = \risk(\vec{\theta}; \data) + \vec{b}^\top \vec{\theta},
  \label{eqn:cr-risk-perturb}
\end{equation}
where $\vec{b} \in \reals^d$ is a vector of coefficients drawn i.i.d.\ from a 
standard normal distribution with standard deviation $\sigma$.
The resulting perturbed learning problem can be solved using standard 
convex optimization methods.
This procedure is summarized in pseudocode below.

\begin{algorithm}
  \begin{algorithmic}[1]
    \Procedure{$\learn$}{$D$} %
      \State $\vec{b} \sim \operatorname{Normal}(\vec{0}, \sigma \cdot \mathbb{I}_d)$
      \State $\vec{\theta}^\star \gets \arg \min_{\vec{\theta}} R_{\vec{b}}(\vec{\theta}; D)$
      \State $\beta \gets 0$ \Comment{initial gradient residual norm is zero}
      \State \textbf{return} $\vec{\theta}^\star, \beta$
    \EndProcedure
  \end{algorithmic}
  \caption{Learning algorithm for certified removal}
  \label{alg:learning}
\end{algorithm}

\paragraph{Unlearning algorithm.}
\label{sec:cr-unlearning}
Approximate unlearning is done by perturbing the 
model parameters $\vec{\theta}$ towards the optimizer of the new objective 
$\risk_{\vec{b}}(\vec{\theta}; \data \setminus \data')$. 
The perturbation is computed as 
\begin{equation*}
  \Delta \vec{\theta} = - \infl(\data'; \data, \vec{\theta})
\end{equation*}
where $\data$ is the initial training data, $\data' \subset \data$ is the 
data to erase, $\vec{\theta}$ are the initial model parameters and
\begin{equation}
  \infl(\data'; \data, \vec{\theta}) = 
    - \left[\hessian_{\vec{\theta}} \risk_{\vec{b}}(\vec{\theta}; \data)\right]^{-1} 
      \nabla_{\vec{\theta}} \risk(\vec{\theta}; \data')
  \label{eqn:influence-fn}
\end{equation}
is the \emph{influence function} \citep{cook1980characterizations,
koh2017understanding}.
The influence function approximates the change in $\vec{\theta}$ when $\data$ is 
augmented by $\data'$.\footnote{
  The influence function is derived assuming $\vec{\theta}$ is the optimizer 
  $\vec{\theta}^\star = \arg \min_{\vec{\theta}} \risk_{\vec{b}}(\vec{\theta}; \data)$. 
  We include $\vec{\theta}$ as an argument since deviations from $\vec{\theta}^\star$ may 
  occur during unlearning. 
  Note that the \emph{negative} influence function captures the scenario 
  where $\data'$ is \emph{subtracted} from $\data$.
}
The first factor in \eqref{eqn:influence-fn} is the inverse Hessian of the 
objective evaluated on $\data$ and the second factor is the gradient of the 
objective evaluated on $\data'$.

Since the update $\vec{\theta} \gets \vec{\theta} + \Delta \vec{\theta}$ is approximate, it is 
not guaranteed to minimize the new objective exactly. 
This becomes a problem for indistinguishability if the error in $\vec{\theta}$ is 
so large that it is not masked by the noise added during learning. 
To ensure indistinguishability, the unlearning algorithm 
maintains an upper bound on the gradient residual norm (GRN) 
$\| \nabla \risk_{\vec{b}}(\vec{\theta} + \Delta \vec{\theta}; \data \setminus \data') \|_2$ 
and resorts to retraining from scratch if a trigger value is exceeded.
Given an initial upper bound on the GRN of $\beta$ (before $\data'$ is 
removed and $\vec{\theta}$ is updated), the upper bound is updated as
\begin{equation}
  \beta \gets \beta + \gamma \|\mat{X}\|_2 \| \Delta \vec{\theta} \|_2 \|\mat{X} \Delta \vec{\theta}\|_2
  \label{eqn:grnb}
\end{equation}
where $\gamma$ is the Lipschitz constant of $\ell''$, $\mat{X}$ is the feature 
matrix associated with $\data \setminus \data'$ and the feature 
vectors are assumed to satisfy $\|\vec{x}\|_2 \leq 1$ for all 
$(\vec{x}, y) \in \data$. 
Whenever $\beta$ exceeds the trigger value
\begin{equation}
  \beta_\mathrm{trigger} = \sigma \epsilon / \sqrt{2 \log 3/2 \delta}
  \label{eqn:beta-trigger}
\end{equation}
the algorithm resorts to retraining from scratch to ensure certified removal, 
as proved by \citet{guo2020certified}.

The entire procedure is summarized in pseudocode below.

\begin{algorithm}
  \begin{algorithmic}[1]
    \Procedure{$\unlearn$}{$D, D', h$} %
      \State $\vec{\theta}, \beta \gets h$
      \State $\Delta \vec{\theta} \gets - \infl(\data'; \data, \vec{\theta})$
      \State $\data \gets \data \setminus \data'$
      \State $\mat{X} \gets$ feature matrix of $D$
      \State $\beta \gets \beta + \gamma \| \mat{X} \|_2 \| \Delta \vec{\theta} \|_2 \| \mat{X} \Delta \vec{\theta} \|_2$
      \If {$\beta > \beta_\mathrm{trigger}$}
        \State $\vec{\theta}, \beta \gets \learn(\data)$ \Comment{Retrain from scratch (slow)}  \label{alg-ln:unlearn-retrain}
      \Else
        \State $\vec{\theta} \gets \vec{\theta} + \Delta \vec{\theta}$ \Comment{Approximate update (fast)}  \label{alg-ln:unlearn-approx}
      \EndIf
      \State \textbf{return} $\vec{\theta}, \beta$
    \EndProcedure
  \end{algorithmic}
  \caption{Unlearning algorithm for certified removal}
  \label{alg:unlearning}
\end{algorithm}

\subsection{Vulnerability and threat model}
\label{sec:threat-model}
One critique of the unlearning algorithm introduced in the previous section
is that it does not guarantee a reduction in computation cost compared to 
full retraining.
In the worst case, the error of the approximate unlearning update exceeds 
the allowed threshold ($\beta > \beta_\mathrm{trigger}$) and the algorithm 
resorts to full retraining (line~\ref{alg-ln:unlearn-retrain} in 
\alref{alg:unlearning}).
We exploit this vulnerability in this paper to design an attack that 
triggers retraining more frequently, thereby diminishing (or even nullifying) 
the promised efficiency gains of unlearning.

\paragraph{Capabilities of the attacker.}
We assume the attacker controls one or more users, who are able to modify (within limits)
their training data and initiate erasure requests. 
In practical settings, the attacker could increase the number of users under 
their control by creating fake user accounts.
Following standard threat models in the data poisoning literature 
\citep{biggio2012poisoning, munoz2017towards, shafahi2018poison, 
shen2019tensorclog, huang2020metapoison} we consider attackers with 
\emph{white-box} and \emph{grey-box} access. 
In our white-box setting, the attacker can access training data of benign 
users, the architecture of the deployed model and the model state 
(excluding the coefficients $\vec{b}$ of the random perturbation term).
While this is a perhaps overly pessimistic view, it allows for a worst-case
evaluation of the attack.
In our grey-box setting, the attacker still has knowledge of the model 
architecture, but can no longer access training data of benign users nor the 
model state.
In place of real training data, the grey-box attacker acquires surrogate data 
from the same or a similar distribution.

\paragraph{Capabilities of the defender.}
For the majority of this paper, we assume the organization (the defender) 
is unaware that unlearning is vulnerable to slow-down attacks and does 
not mount any specific defenses.
However, we do assume the organization conducts basic data validity checks
(e.g.\ ensuring image pixel values are within valid ranges) and occasional 
manual auditing of data. 
To evade detection during manual auditing, we assume the attacker makes 
(small) bounded modifications to clean data which are unlikely to arouse 
suspicion.

\section{Slow-down Attack}
\label{sec:attack}

In this section, we present an attack on unlearning algorithms with 
data-dependent running times. 
We begin by formulating our attack as a generalized data poisoning problem 
in \secref{sec:bilevel-opt}.
Then in \secref{sec:pgd}, we describe a practical solution based on 
projected gradient descent under some simplifying assumptions.
Finally, in \secref{sec:cost-fn} we propose concrete objectives 
for the attacker and suggest further relaxations to speed-up the attack.

\subsection{Formulation as a data poisoning problem}
\label{sec:bilevel-opt}
Let $\data$ be data used by the organization (the defender) to train (with 
learning algorithm \(A\)) an initial model \(\hat h = \learn(\data)\).
Suppose the attacker is able to poison a subset $\data_\pois$ of 
instances in $\data$, and denote the remaining clean instances by 
$\data_\clean = \data \setminus \data_\pois$.
We introduce a function $C: \hyposet, \dataspace \to \reals$ which measures 
the \emph{computational cost} $C(\hat{h}, \data')$ of erasing a subset of 
data $\data' \subset \data$ from trained model $\hat{h} \in \hyposet$ using 
unlearning algorithm $\unlearn$.

The attacker's aim is to craft $\data_\pois$ 
to maximize $C(\hat{h}, \data_\pois)$.
Assuming the attacker enforces constraints on $\data_\pois$ to pass the 
defender's checks\slash auditing, we express the attacker's problem 
formally as follows:
\begin{subequations}
    \begin{alignat}{2}
        &\!\max_{\data_\pois} &\qquad& C(\hat{h}, \data_\pois) \label{eqn:bilevel-upper}\\
        &\text{subject to}    &      & \hat{h} = \E[\learn(\data_\clean \cup \data_\pois)] \label{eqn:bilevel-lower}\\
        &                     &      & g_j(\data_\pois) \leq 0 \quad \forall j \in \{1, \ldots, J\} \label{eqn:bilevel-inequal}
    \end{alignat}
\end{subequations}
This is a \emph{bilevel optimization problem} \citep{mei2015using} if the 
learning algorithm $\learn$ solves a deterministic optimization problem. 
Although solving \eqref{eqn:bilevel-upper}--\eqref{eqn:bilevel-inequal} 
exactly is infeasible in general, we can find locally-optimal solutions 
using gradient-based methods, which we discuss next.

\subsection{PGD-based crafting strategy}
\label{sec:pgd}
We now outline an approximate solution to 
\eqref{eqn:bilevel-upper}--\eqref{eqn:bilevel-inequal} based on projected 
gradient descent (PGD).
Our solution makes the following assumptions:
\begin{enumerate}
  \item The learning\slash unlearning algorithms adopted by the organization 
  (defender) are those presented in \secref{sec:cr} for regularized 
  linear models (Algs.~\ref{alg:learning} and~\ref{alg:unlearning}). 
  \item The attacker crafts the poisoned data $\data_\pois$ using clean 
  reference data $\data_\mathrm{ref}$ as a basis. 
  In addition, we assume the attacker only modifies features in the 
  reference data, leaving labels unchanged.
  Mathematically, if we write $\data_\pois = (\mat{X}_\pois, \vec{y}_\pois)$ 
  and $\data_\mathrm{ref} = (\mat{X}_\mathrm{ref}, \vec{y}_\mathrm{ref})$ where 
  $\mat{X}_\pois, \mat{X}_\mathrm{ref} \in \reals^{m \times d}$ are feature 
  matrices and $\vec{y}_\pois, \vec{y}_\mathrm{ref} \in \reals^m$ are label 
  vectors, then the attacker optimizes $\mat{X}_\pois$ and fixes 
  $\vec{y}_\pois = \vec{y}_\mathrm{ref}$. 
  \item The inequality constraints in \eqref{eqn:bilevel-inequal} are 
  of the form 
  \begin{equation}
    g_j(\data_\pois) = \sup_{\vec{x} \in \mathrm{rows}(\mat{X}_\pois - \mat{V}_j)} 
      \left\| \vec{x} \right\|_{p_j} - r_j
    \label{eqn:inequal-p-ball}
  \end{equation}
  where $\vec{x} \in \reals^{1 \times d}$, $\mat{V}_j \in \reals^{m \times d}$, 
  $r_j > 0$ and $p_j \in \mathbb{Z}_{>0}$. 
  This ensures each row of $\mat{X}_\pois$ is confined to an $\ell_{p_j}$-ball 
  of radius $r_j$ centered on the corresponding row in $\mat{V}_j$.
\end{enumerate}

In addition to these assumptions, we approximate the expectation in 
\eqref{eqn:bilevel-lower} since it cannot be computed exactly.
Applying a zero-th order expansion of the $\arg\max$ function, we make the 
replacement:
\begin{equation*}
  \E[\learn(\data)] \to \arg \max_{\vec{\theta}} \E[\risk_{\vec{b}}(\vec{\theta}; \data)] 
                = \arg \max_{\vec{\theta}} \risk(\vec{\theta}; \data).
\end{equation*}
The equality follows from the definition of the perturbed objective in 
\eqref{eqn:cr-risk-perturb} and the fact that the expectation of the 
perturbation is zero.

Putting everything together, we recast \eqref{eqn:bilevel-upper}--\eqref{eqn:bilevel-inequal}
as a single-level constrained optimization problem:
\begin{subequations}
  \begin{alignat}{2}
      &\!\min_{\mat{X}_\pois \in \reals^{m \times d}} &\qquad& f(\mat{X}_\pois) \label{eqn:singlelevel} \\ 
      &                                               &      & g_j(\data_\pois) \leq 0 \quad \forall j \in \{1, \ldots, J\} \label{eqn:singlelevel-inequal}
  \end{alignat}
\end{subequations}
where we define 
\begin{gather}
  f(\mat{X}_\pois) = -C(\hat{\vec{\theta}}(\data), \data_\pois) \text{ and } 
    \hat{\vec{\theta}}(\data) = \arg \min_{\vec{\theta}} \risk(\vec{\theta}; \data).
    \label{eqn:simple-obj}
\end{gather}
This problem can be solved using projected gradient descent (PGD) as detailed 
in Appendix~\ref{app:pgd}.

\subsection{Measuring the computational cost}
\label{sec:cost-fn}
Recall that the attacker's goal is to maximize the computational cost 
$C(\hat{\vec{\theta}}, \data_\pois)$ of erasing poisoned data $\data_\pois$ from 
the model $\hat{\vec{\theta}}$ trained on $\data = \data_\pois \cup \data_\clean$.
In this section, we discuss practical choices for $C$ assuming the 
organization  implements the learning\slash unlearning algorithms 
in \secref{sec:cr}.

We begin with the observation that the computational cost of erasing data 
in a call to $\unlearn$ (\alref{alg:unlearning}) varies considerably 
depending on whether full retraining is triggered or not. 
\citet{guo2020certified} report that full retraining is three 
orders of magnitude slower than an approximate update, making it the 
dominant contribution when averaged over a series of calls to $\unlearn$.
Complete retraining is triggered when the gradient residual norm bound 
$\beta$ exceeds a threshold, and therefore we model the attacker as aiming to 
maximize $\beta$.

Using the expression for $\beta$ in \eqref{eqn:grnb}, we therefore set
\begin{equation}
  C(\hat{\vec{\theta}}, \data_\pois) = \| \mat{X} \|_2 \| \Delta \vec{\theta} \|_2 \| \mat{X} \Delta \vec{\theta} \|_2
  \label{eqn:cost-grnb}
\end{equation}
where $\mat{X}$ is the feature matrix associated with 
$\data \setminus \data_\pois$ and 
$\Delta \vec{\theta} = - \infl(\data_\pois; \data, \hat{\vec{\theta}})$.\footnote{
  For readability, we omit factors that are independent of $\data$ when
  defining cost functions.
}
We note that this cost function assumes $\data_\pois$ is erased in a 
\emph{single call} to $\unlearn$.
However our experiments in \secref{sec:experiments} indicate that 
\eqref{eqn:cost-grnb} is effective even if $\data_\pois$ is erased in a 
\emph{sequence of calls} to $\unlearn$.

\paragraph{Faster cost surrogates.}
While \eqref{eqn:cost-grnb} governs whether retraining is triggered, it may be 
expensive for the attacker to evaluate, particularly due to the presence of 
$\mat{X}$ which requires operations that scale linearly in $n - m$ (the size of 
the remaining training data).
We therefore consider simpler surrogates for the computational cost, based on 
alternate upper bounds for the gradient residual norm (GRN).
We study the effect of these surrogates empirically in 
\secref{sec:experiments}.

The first surrogate is based on a data-independent upper bound of the GRN from 
\citep[Theorem~1]{guo2020certified}:
\begin{equation}
  C(\hat{\vec{\theta}}, \data_\pois) = \| \infl(\data_\pois; \data, \hat{\vec{\theta}}) \|_2.
  \label{eqn:cost-infl}
\end{equation}
Since this is a looser bound on the GRN than \eqref{eqn:cost-grnb}, we expect 
it will produce less effective attacks.

The second surrogate upper bounds \eqref{eqn:cost-infl}. 
Observing that 
\begin{align*}
  \|\infl(\data_\pois; \data, \hat{\vec{\theta}})\|_2 
    & \leq \|\hessian_{\vec{\theta}} \risk_{\vec{b}}(\hat{\vec{\theta}}; \data)\|_2 
      \cdot \| \nabla_{\vec{\theta}} \risk(\hat{\vec{\theta}}; \data_\pois) \|_2 \\
    & \leq \frac{1}{\lambda(|\data| - 1)} \cdot \| \nabla_{\vec{\theta}} \risk(\hat{\vec{\theta}}; \data_\pois) \|_2 
\end{align*}
we propose
\begin{equation}
  C(\hat{\vec{\theta}}, \data_\pois) = \| \nabla_{\vec{\theta}} \risk(\hat{\vec{\theta}}; \data_\pois) \|_2.
  \label{eqn:cost-grad}
\end{equation}
This doesn't depend on the Hessian (given $\hat{\vec{\theta}}$) and is therefore 
significantly cheaper for the attacker to compute.

\paragraph{Ignoring model dependence.}
Another way of reducing the attacker's computational effort is to assume 
the model trained on $\data = \data_\pois \cup \data_\clean$ is 
approximately independent of $\data_\pois$. 
Specifically, we make the replacement 
$\hat{\vec{\theta}} = \arg \max_{\vec{\theta}} \risk(\vec{\theta}; \data_\clean)$ in 
\eqref{eqn:simple-obj} so that $\hat{\vec{\theta}}$ is constant with respect to 
$\data_\pois$. 
This means evaluating the objective (or its gradient) no longer requires 
retraining and simplifies gradient computations (see Appendix~\ref{app:pgd}).
Although this approximation is based on a dubious assumption---the model 
must be somewhat sensitive to $\data_\pois$---we find it performs 
well empirically (see Table~\ref{tbl:retrain-int-surrogate}).

\section{Experiments}
\label{sec:experiments}

We investigate the impact of our attack on the computational cost of 
unlearning in a variety of simulated settings. 
We study the effect of the organization's parameters---the regularization 
strength $\lambda$ and the magnitude of the objective perturbation 
$\sigma$---as well as the attacker's parameters---the cost function $C$, 
and the type of norm and radius $r$ used to bound the perturbations.
We further evaluate the effectiveness of our attack over time, as additional 
poisoned erasure requests are processed.
Finally, we investigate a transfer setting where 
the attacker uses a surrogate model (trained from surrogate data) in lieu 
of the defender's true model.
Code is available at \url{https://github.com/ngmarchant/attack-unlearning}.

\subsection{Datasets and setup}
\label{sec:expt-setup}
We consider MNIST \citep{lecun1998gradient} and Fashion-MNIST 
\citep{xiao2017fashionmnist}, both of which contain $d = 28 \times 28$ 
single-channel images from ten classes. 
We also generate a smaller binary classification dataset from MNIST we call 
Binary-MNIST which contains classes 3 and 8.
Beyond the image domain, we consider human activity recognition (HAR) data 
\citep{anguita2013public}.
It contains windows of processed sensor signals ($d = 561$) corresponding to 
6 activity classes. 

All datasets come with predefined train\slash test splits.
The way the splits are used varies for each attack setting (defined  
in \secref{sec:threat-model}).
In the white-box setting, the attacker accesses the train split and 
\emph{replaces} instances under their control\footnote{%
  The instances under the attackers control are randomly-selected in each trial.
} with poisoned instances. 
The test split is used to estimate model accuracy.
In the grey-box setting, the train split is inaccessible to the 
attacker.
Instead the attacker uses the test split as surrogate data to craft poisoned 
instances, which are then \emph{added} to the initial train split.
We do not report model accuracy for this setting.
In both settings, the resulting training data $\data$ is used by the 
organization to learn a logistic regression model, after which the attacker 
submits erasure requests for the poisoned instances $\data_\pois \subset \data$ 
one-by-one.

The main quantity of interest is the \emph{retrain interval}---the number 
of erasure requests handled via fast approximate updates before full 
retraining is triggered. 
A more effective attack achieves a smaller retrain interval.
Further details about the experiments, including hardware, parameter settings 
poisoning ratios, and the number of trials in each experiment are provided in 
Appendix~\ref{app:expt-detail}.

\subsection{Results}
\label{sec:expt-results}

\paragraph{Varying model sensitivity.} 
Table~\ref{tbl:retrain-int-sigma-lamb} shows that our attack can reduce 
the retrain interval by 70--100\% over a range of settings for $\lambda$ and 
$\sigma$.
The defender can adjust the model's sensitivity to training data by 
varying the regularization strength $\lambda$ and the magnitude of the 
objective perturbation $\sigma$. 
Our attack is most effective in an absolute sense for smaller values of 
$\lambda$ and $\sigma$, where the retrain interval is reduced to zero. 
This means retraining is triggered immediately upon processing a poisoned 
erasure request. 
Table~\ref{tbl:retrain-int-sigma-lamb} also illustrates a trade-off between 
model accuracy and unlearning efficiency: larger values of $\lambda$ 
and $\sigma$ generally allow for more efficient unlearning (larger retrain 
interval) at the expense of model accuracy.

\begin{table}[t]
  \caption{Attack effectiveness on Binary-MNIST as a function of the 
  regularization strength $\lambda$ and magnitude of the objective 
  perturbation $\sigma$.
  The accuracy is reported for the initial model, prior to processing 
  erasure requests.}
  \label{tbl:retrain-int-sigma-lamb}
  \centering
  \ifarxiv \else \footnotesize \fi 
  \begin{tabularx}{\ifarxiv 0.75\linewidth \else \linewidth \fi}{ccRRRRr}
    \toprule
    $\sigma$ & $\lambda$ & \multicolumn{2}{c}{Accuracy} & \multicolumn{3}{c}{Retrain interval} \\
    \cmidrule(lr){3-4} \cmidrule(lr){5-7}
             &           &       Benign &        Attack &      Benign &      Attack &  \% $\downarrow$ \\
    \midrule
    \multirow{4}{*}{1} 
             & $10^{-5}$ &        0.962 &         0.962 &        3.58 &        0.07 &     98.0 \\
             & $10^{-4}$ &        0.968 &         0.968 &        5.80 &           0 &      100 \\
             & $10^{-3}$ &        0.958 &         0.959 &        16.4 &        0.22 &     98.7 \\
             & $10^{-2}$ &        0.926 &         0.926 &         188 &        48.4 &     74.3 \\
    \midrule
    \multirow{4}{*}{10}
             & $10^{-5}$ &        0.919 &         0.919 &           0 &           0 &       -- \\
             & $10^{-4}$ &        0.932 &         0.931 &        9.32 &        0.27 &     97.1 \\
             & $10^{-3}$ &        0.954 &         0.955 &         132 &        8.15 &     93.8 \\
             & $10^{-2}$ &        0.926 &         0.920 &        1640 &         512 &     68.8 \\
    \bottomrule
  \end{tabularx}
\end{table}

\paragraph{Varying the peturbation constraint.}
Table~\ref{tbl:retrain-int-radius} shows how the effectiveness of our attack 
varies depending on the choice of $\ell_p$-norm and radius $r$. 
As expected, the attack is most effective when there is no perturbation 
constraint ($p = \infty$ and $r = 1$ or $p = 1$ and $r = d$), however 
the poisoned images are no longer recognizable as digits. 
The $\ell_1$-norm constraint appears more effective, as it can better exploit 
sensitivities in the weights of individual pixels.

\begin{table}[t]
  \newlength{\imwidth}
  \setlength{\imwidth}{0.66cm}
  \caption{Attack effectiveness on Binary-MNIST as a function of the 
    perturbation constraint.}
  \label{tbl:retrain-int-radius}
  \centering
  \ifarxiv \else \footnotesize \fi
  \begin{tabular}{ccrc}
    \toprule
    \multicolumn{2}{c}{Constraint} &  Retrain interval & Poisoned examples \\
    \cmidrule(lr){1-2}
    Norm &   $r$ &                              \\
    \midrule
    \multirow{5}{*}[-14pt]{$\ell_1$} 
         &        0 &                                131 & 
         \raisebox{-.2\height}{%
         \includegraphics[width=\imwidth]{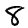} 
         \includegraphics[width=\imwidth]{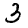} 
         \includegraphics[width=\imwidth]{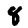} 
         \includegraphics[width=\imwidth]{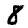}} \\[2pt]
         &  $d/200$ &                               72.3 & 
         \raisebox{-0.2\height}{%
         \includegraphics[width=\imwidth]{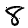} 
         \includegraphics[width=\imwidth]{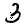} 
         \includegraphics[width=\imwidth]{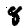} 
         \includegraphics[width=\imwidth]{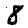}} \\[2pt]
         &   $d/20$ &                               8.15 & 
         \raisebox{-.2\height}{%
         \includegraphics[width=\imwidth]{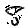} 
         \includegraphics[width=\imwidth]{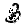} 
         \includegraphics[width=\imwidth]{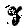} 
         \includegraphics[width=\imwidth]{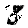}} \\[2pt]
         &    $d/2$ &                               3.42 &
         \raisebox{-.2\height}{%
         \includegraphics[width=\imwidth]{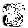} 
         \includegraphics[width=\imwidth]{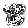} 
         \includegraphics[width=\imwidth]{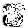} 
         \includegraphics[width=\imwidth]{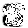}} \\[2pt]
         &      $d$ &                               3.54 &
         \raisebox{-.2\height}{%
         \includegraphics[width=\imwidth]{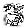} 
         \includegraphics[width=\imwidth]{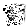} 
         \includegraphics[width=\imwidth]{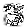} 
         \includegraphics[width=\imwidth]{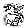}} \\[2pt]
    \midrule
    \multirow{5}{*}[-14pt]{$\ell_\infty$}  
         &    0.000 &                                131 &
         \raisebox{-.2\height}{%
         \includegraphics[width=\imwidth]{mnist_binary_0.png} 
         \includegraphics[width=\imwidth]{mnist_binary_1.png} 
         \includegraphics[width=\imwidth]{mnist_binary_2.png} 
         \includegraphics[width=\imwidth]{mnist_binary_3.png}} \\[2pt]
         &    0.050 &                               82.2 &
         \raisebox{-.2\height}{%
         \includegraphics[width=\imwidth]{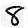} 
         \includegraphics[width=\imwidth]{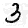} 
         \includegraphics[width=\imwidth]{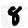} 
         \includegraphics[width=\imwidth]{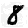}} \\[2pt]
         &    0.100 &                               48.9 &
         \raisebox{-.2\height}{%
         \includegraphics[width=\imwidth]{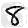} 
         \includegraphics[width=\imwidth]{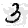} 
         \includegraphics[width=\imwidth]{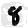} 
         \includegraphics[width=\imwidth]{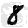}} \\[2pt]
         &    0.500 &                               5.63 &
         \raisebox{-.2\height}{%
         \includegraphics[width=\imwidth]{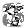} 
         \includegraphics[width=\imwidth]{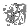} 
         \includegraphics[width=\imwidth]{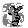} 
         \includegraphics[width=\imwidth]{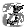}} \\[2pt]
         &    1.000 &                               3.54 &
         \raisebox{-.2\height}{%
         \includegraphics[width=\imwidth]{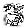} 
         \includegraphics[width=\imwidth]{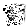} 
         \includegraphics[width=\imwidth]{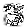} 
         \includegraphics[width=\imwidth]{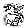}} \\[2pt]
    \bottomrule
  \end{tabular}
\end{table}

\paragraph{Varying the cost function.}
We study whether the attack remains effective when: (i)~faster surrogate cost 
functions are used and (ii)~the model dependence on the poisoned 
data $\data_\pois$ is ignored (see \secref{sec:cost-fn}).
Table~\ref{tbl:retrain-int-surrogate} shows the attack can be executed with 
a similar effectiveness (retrain interval) using the influence norm cost 
function \eqref{eqn:cost-infl} in place of the GRNB cost function 
\eqref{eqn:cost-grnb}, while reducing the attacker's computation. 
The gradient norm cost function \eqref{eqn:cost-grad} is the least 
computationally demanding for the attacker, but less effective. 
We see that it is reasonable to ignore the model dependence on $\data_\pois$, 
as there is no marked difference in effectiveness and a significant reduction 
in the attacker's computation. 
We expect this is due to the relatively small size of $\data_\pois$ 
(reflecting realistic capabilities of an attacker) which limits the impact 
of poisoning on the model. 

\begin{table}[t]
  \caption{Attack effectiveness and computation time for different choices of 
     the cost function.}
  \label{tbl:retrain-int-surrogate}
  \centering
  \ifarxiv \else \footnotesize \fi
  \begin{tabularx}{\ifarxiv 0.75\linewidth \else \linewidth \fi}{lLRR}
    \toprule
    Cost function & Ignore model dep. & Retrain interval & Attack time (s) \\
    \midrule
    \multirow{2}{*}{GRNB \eqref{eqn:cost-grnb}}      
              &                     No  &             6.96 &            39.2 \\
              &                     Yes &             7.08 &            24.4 \\
    \midrule
    \multirow{2}{*}{Influence norm \eqref{eqn:cost-infl}} 
              &                      No &             7.98 &            29.0 \\
              &                     Yes &             8.15 &            8.72 \\
    \midrule
    \multirow{2}{*}{Gradient norm \eqref{eqn:cost-grad}}
              &                      No &            16.34 &            23.1 \\
              &                     Yes &            19.21 &            3.54 \\
    \bottomrule 
  \end{tabularx}
\end{table}

\begin{figure}[t]
  \centering
  \includegraphics[width=\ifarxiv 0.60\linewidth \else 0.95\linewidth \fi]%
  {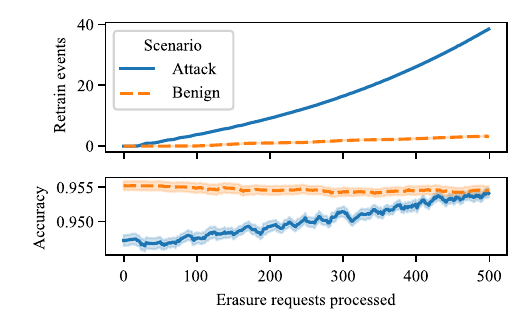}
  \caption{Number of times retraining is triggered as a function of the number 
    of poisoned erasure requests processed.
    After processing 500~requests, all poisoned instances are erased and only 
    clean data remains.}
  \label{fig:mnist-binary-prolonged-attack}
\end{figure}

\paragraph{Long-term effectiveness.}
We are interested in whether the attack effectiveness varies over time---as 
poisoned erasure requests are processed. 
Figure~\ref{fig:mnist-binary-prolonged-attack} shows a moderate increase 
in effectiveness, indicated by the increasing rate of retraining. 
The is likely due to the fact that the defender's model is initially far 
from the model used by the attacker (the former is trained on $\data$, the 
latter on $\data_\clean$). 
However as poisoned instances are erased, the defender's model approaches 
the attacker's model, and the poisoned instances become more effective.

\paragraph{Transferability.}
We investigate whether the attack transfers if \emph{surrogate data} is 
used to craft the attack in place of the defender's training data. 
This corresponds to the grey-box setting described in 
\secref{sec:threat-model}.
Table~\ref{tbl:retrain-int-transfer} shows that the attack transfers 
well---there is very little difference in the retrain interval in the 
grey- versus white-box settings.

\begin{table}[t]
  \caption{Effectiveness of the attack in white-box versus grey-box settings.}
  \label{tbl:retrain-int-transfer}
  \centering
  \ifarxiv \else \footnotesize \fi
  \begin{tabularx}{\ifarxiv 0.75\linewidth \else \linewidth \fi}{lRR}
    \toprule
    Dataset      & \multicolumn{2}{c}{Retrain interval} \\
    \cmidrule(lr){2-3}
                 &    Surrogate data (Grey-box) &    Training data (White-box)\\
    \midrule
    MNIST        &              19.0 &             18.5 \\
    Binary-MNIST &              7.88 &             8.15 \\
    \bottomrule
  \end{tabularx}
\end{table}

\paragraph{Varying datasets.} 
Most of the results reported above are for Binary-MNIST. 
Table~\ref{tbl:retrain-int-data} reports results for other datasets 
described in \secref{sec:expt-setup}.
While the results are qualitatively similar across datasets, we note some variance in attack effectiveness---depending on the 
dataset we observe a reduction in the retrain interval of 67--94\%.
The size of each dataset is likely a factor, since a model trained on more 
data is less sensitive to individual instances and more difficult to attack. 
We also expect the $\ell_1$-bounded perturbations are likely to have a 
different impact depending on the characteristics of the data.
For instance, Fashion-MNIST has a wider variation of intensities across 
all pixels compared to MNIST, which may make $\ell_1$-bounded 
perturbations less effective.

\begin{table}[t]
  \caption{Effectiveness of the attack for various datasets.}
  \label{tbl:retrain-int-data}
  \centering
  \ifarxiv \else \footnotesize \fi
  \begin{tabularx}{\ifarxiv 0.75\linewidth \else \linewidth \fi}{lRRRRr}
    \toprule
    Dataset        & \multicolumn{2}{c}{Accuracy} & \multicolumn{3}{c}{Retrain interval} \\
    \cmidrule(lr){2-3} \cmidrule(lr){4-6}
                   &       Benign &        Attack &      Benign &      Attack &  \% $\downarrow$ \\
    \midrule    
    MNIST          &        0.867 &         0.867 &        71.6 &        18.5 &    74.1 \\
    Binary-MNIST   &        0.954 &         0.954 &         132 &        8.15 &    93.8 \\
    Fashion-MNIST  &        0.756 &         0.756 &         119 &        38.9 &    67.5 \\
    HAR            &        0.838 &         0.836 &        42.2 &        8.95 &    78.8 \\
    \bottomrule
  \end{tabularx}
\end{table}

\section{Related Work}
\paragraph{Machine unlearning.}

Recent interest in efficient data erasure for learned models was spawned 
by \citet{cao2015towards}, who were motivated by privacy and security 
applications. 
They coined the term ``machine unlearning'' and proposed a strict 
definition---requiring that an unlearning algorithm return a model 
\emph{identical} to one retrained from scratch on the remaining data. 
Efficient unlearning algorithms are known for some canonical models 
under this definition, including linear regression 
\citep{chambers1971regression, hansen1996linear}, %
naive Bayes and non-parametric models such as $k$-nearest neighbors 
\citep{schelter2020amnesia}. 
These algorithms are immune to our attack since they compute updates in 
closed form with data-independent running times.

A strict definition of unlearning has proven difficult to satisfy for 
general models. 
One solution has been to design new models or adapt existing ones with 
unlearning efficiency in mind. 
Examples of such models include ensembles of randomized 
decision trees \citep{schelter2021hedgecut}, variants of $k$-means 
clustering \citep{ginart2019making}, and a modeling framework called 
SISA that takes advantage of data sharding and caching 
\citep{bourtoule2021machine}.

Others have expanded the scope of unlearning by adopting more relaxed  
(but still rigorous) definitions. 
\citet{ginart2019making} were first to introduce an approximate definition 
of unlearning---requiring that it be \emph{statistically indistinguishable} 
from retraining.
They identified a connection to differential privacy \citep{dwork2006our} 
and suggested adversarial robustness as a direction for future work.
Subsequent work \citep{guo2020certified, neel2021descent, 
sekhari2021remember} has adopted similar approximate definitions to provide 
certified unlearning guarantees for strongly-convex learning problems. 
\citet{neel2021descent} and \citet{sekhari2021remember} study 
asymptotic time complexity, however they do not test their methods 
empirically and they rely on data-independent bounds which are too loose 
in practice according to \citet{guo2020certified}. 
Since the approach by \citeauthor{guo2020certified} is empirically 
validated, we use it for our proof-of-concept exploit in this paper.

Certified unlearning for non-convex models such as deep neural networks 
(DNNs) remains elusive. 
\citet{golatkar2020eternal, golatkar2020forgetting} have made some progress
in this area based on a notion of approximate unlearning. 
However, their methods do not guarantee approximate unlearning is satisfied 
and they require further approximations to scale to DNNs.
Since their methods run for a predetermined number of iterations, the 
running time is data-independent, making them immune to slow-down attacks.
However, our slow-down attack may cause damage in a different way---preventing 
proper data erasure or leaving the unlearned model in an unpredictable state. 

While our focus in this paper is on attacking unlearning efficiency, others 
have considered privacy vulnerabilities.
\citet{chen2020machine} demonstrated that a series of published unlearned 
models are vulnerable to differencing attacks if not appropriately treated.
\citet{gupta2021adaptive} considered an adaptive unlearning setting, where 
erasure requests may depend on previously published models. 
They also demonstrated an attack on the SISA algorithm 
\citep{bourtoule2021machine}. 
\citet{sommer2020probabilistic} considered verification of unlearning---an 
important privacy feature for users who cannot trust organizations to erase 
their data.

\paragraph{Adversarial machine learning.}

Our work builds on standard methods for \emph{data poisoning}---a class of 
attacks that manipulate learned models by modifying or augmenting training 
data (see survey \citealp{schwarzschild2021just}). %
Two commonly studied variants of data poisoning are \emph{availability attacks}, 
which aim to degrade model test performance (e.g.\ 
\citealp{biggio2012poisoning, munoz2017towards, koh2017understanding, 
shafahi2018poison, zhu2019transferable, huang2020metapoison}),  %
and \emph{backdoor attacks}, which aim to cause misclassification of test-time 
samples that contain a trigger (e.g. \citealp{chen2017targeted, dai2019backdoor, 
saha2020hidden}). 
Unlike these attacks, our attack does not aim to harm test performance---rather 
the aim is to cause trouble at the unlearning stage. 
This adds another layer of complexity, as we must account for learning and 
unlearning to craft an effective attack.

Most work on data poisoning assumes learning is done offline, as we do 
in this paper. 
However one could imagine an online scenario, where an organization 
continually updates a model as new data arrives and old data is erased.
There has been some work on data poisoning for online learning 
\citep{wang2018data, zhang2020online}, however existing  
work does not account for the ability of an attacker to request that their data be erased.

Our attack also borrows ideas from the adversarial examples literature. 
The $\ell_p$-norm constraints that we impose on the attacker's 
perturbations are commonly used to generate adversarial examples 
\citep{szegedy2014intriguing, goodfellow2015explaining}. 
While $\ell_p$-norm constraints are easy to work with, they have been 
criticized for failing to capture perceptual distortion 
\citep{sharif2018suitability}, which has lead to work on more effective 
perturbation constraints \citep{zhao2020towards, ballet2019imperceptible}.
These could be leveraged to enhance our attack.

While adversarial examples are predominantly used for evasion attacks, 
they have also been deployed for other purposes. 
\citet{hong2021panda} showed that techniques for generating adversarial 
examples can be modified to slow down inference of multi-exit DNNs by a 
factor of 1.5--5$\times$. 
This attack is qualitatively similar to ours, in that it targets computation 
cost, however the formulation is different. 
While our attack operates at training-time and unlearning-time, the 
attack by \citeauthor{hong2021panda} operates at test-time. 
Other novel applications of adversarial examples include data poisoning 
\citep{tao2021provable} and protecting data from being used to train 
models \citep{huang2021unlearnable}.

\section{Discussion}
We have demonstrated a broad range of settings where a poisoning attack can be 
successfully launched against state-of-the-art machine unlearning to 
significantly undo the advantages of unlearning over full retraining. 
Our results suggest that theory should incorporate a trade-off between 
indistinguishability, computational cost, and accuracy. 
Indeed in parallel work to our own, \citet{neel2021descent} define ``strong'' 
unlearning algorithms as those for which accuracy bounds are independent of 
the number of unlearning requests $t$, with a computation cost that grows at 
most logarithmically in $t$. 
They explore trade-offs for an unlearning algorithm called perturbed gradient 
descent. 
However this approach hasn't been evaluated empirically to our knowledge, and 
a number of open problems remain. 

Future work could apply or extend our methods to attack other unlearning 
algorithms, beyond those by \citet{guo2020certified}, which were the focus 
of this paper. 
We expect this would be easiest for algorithms based on a similar 
design---e.g.\ algorithms by \citet{neel2021descent} and \citet{sekhari2021remember} 
also perturb the original model to find an approximate solution to a  
strongly-convex learning objective post-erasure. 
Other unlearning algorithms, such as \citeauthor{ginart2019making}'s 
algorithm for k-means clustering, are also vulnerable to slow-down attacks, 
however crafting poisons to trigger retraining would be more difficult due to 
non-differentiability of the cost function. 
If certified unlearning algorithms are developed for deep models in the 
future, then assessing the impact of slow-down attacks in that setting---where 
data and models are typically at a much larger scale---would also be of 
immense interest.

Another direction suggested by our work is counter measures against 
slow-down attacks on unlearning. 
One might adopt an unlearning algorithm that never resorts to retraining 
from scratch~\citep{golatkar2020eternal}. 
Such an algorithm might still require more computational effort to remove 
poisoned examples than benign examples, exposing a similar attack surface. 
It is also unclear whether such an algorithm could maintain a form of 
indistinguishability indefinitely. 

One might seek to filter out poisoned examples so they never need to be 
unlearned. 
This might involve off-the-shelf anomaly detection methods, however poisoned 
examples are notoriously difficult to detect when they are crafted as 
small perturbations to clean examples. 
\citet{metzen2017detecting} for instance, have explored detection of 
adversarial perturbations. 
Alternatively, one could filter out examples with a large influence (which the 
attacker is trying to artificially inflate in our attack), however some clean 
examples have a naturally large influence. 

Another approach would be a robust model that is less sensitive to individual 
examples. 
Directions for achieving this include: increasing the degree of 
regularization, introducing noise,  adversarial training, and using more 
training data.
However, like their analogues in existing adversarial learning research, 
these mitigations tend to harm accuracy.

\section*{Acknowledgment}

This research was undertaken using the HPC-GPGPU Facility hosted at the 
University of Melbourne, established with the assistance of ARC LIEF Grant 
LE170100200. 
This research was also supported in part by the Australian Department of 
Defence Next Generation Technologies Fund CSIRO/Data61 CRP AMLC project.

\printbibliography

\clearpage
\appendix

\section{Technical details for PGD}
\label{app:pgd}

In this appendix, we present algorithms and discuss technical considerations 
for our PGD-based solution to the optimization problem 
in \eqref{eqn:singlelevel}--\eqref{eqn:singlelevel-inequal}. 
Our implementation of PGD is summarized in \alref{alg:pgd}, which 
calls sub-routines for backtracking line search (\alref{alg:armijo}) 
and the projection operator (\alref{alg:proj}).

\begin{algorithm}
  \begin{algorithmic}[1]
      \Procedure{PGD}{$\mat{X}_0$}
      \State $\mat{X}_\pois \gets X_0$
      \For {$i \in \{1, \ldots, n_\mathrm{pgd}\}$}
          \State $\mat{G} \gets \nabla f(\mat{X}_\pois)$ \Comment{Compute gradient} \label{alg-ln:pgd-grad}
          \State $\Delta \mat{X}_\pois \gets \begin{bmatrix} 
              \frac{G_{1\cdot}}{\|G_{1\cdot}\|_2} \\ 
              \vdots \\ 
              \frac{G_{m\cdot}}{\|G_{m\cdot}\|_2} 
            \end{bmatrix}$ 
            \Comment{Normalize gradient} \label{alg-ln:pgd-norm-grad}
          \State $\eta \gets \textsc{BLS}(\eta_0, \mat{X}_\pois, \Delta \mat{X}_\pois, \mat{G})$ \Comment{Get step size} \label{alg-ln:pgd-bls}
          \State $\mat{X}_\pois \gets \mat{X}_\pois - \eta \Delta \mat{X}_\pois$ \Comment{Descent step} \label{alg-ln:pgd-proj}
          \State $\mat{X}_\pois \gets \textsc{Proj}(\mat{X}_\pois)$ \Comment{Project step}
      \EndFor
      \State \textbf{return} $\mat{X}_\pois$
      \EndProcedure
  \end{algorithmic}
  \caption{Projected gradient descent with backtracking line search and 
    gradient normalization.}
  \label{alg:pgd}
\end{algorithm}  

\subsection{Computing gradients}
In line~\ref{alg-ln:pgd-grad} of \alref{alg:pgd} we compute the gradient 
of the objective $f$ with respect to the poisoned features $\mat{X}_\pois$.
This is nontrivial since we need to propagate gradients through 
$\hat{\vec{\theta}}(\data)$, which is implemented as an iterative optimization 
algorithm.
Na\"ively applying automatic differentiation to compute 
$\partial \hat{\vec{\theta}}/\partial \mat{X}_\pois$ would require unrolling 
loops in the optimization algorithm.
Fortunately, we can tackle the computation more efficiently using the 
implicit function theorem \citep[Proposition A.25]{bertsekas1999nonlinear}.
Assuming the optimization algorithm runs to convergence, the solution 
$\hat{\vec{\theta}}$ satisfies $\nabla_{\vec{\theta}} \risk(\hat{\vec{\theta}}; \data) = 0$. 
Applying the implicit function theorem on 
$q(\hat{\vec{\theta}}, \mat{X}_\pois) = \nabla_{\vec{\theta}} \risk(\hat{\vec{\theta}}; \data)$ 
gives
\begin{equation}
  \frac{\partial \hat{\vec{\theta}}}{\partial \mat{X}_\pois} = 
    - [\jac_{\vec{\theta}} q(\hat{\vec{\theta}, \mat{X}_\pois})]^{-1} 
      \cdot \jac_{\mat{X}_\pois} q(\hat{\vec{\theta}}, \mat{X}_\pois)
  \label{eqn:custom-grad}
\end{equation}
where $\jac$ is the Jacobian operator.
This expression be used to implement a custom gradient for $\hat{\vec{\theta}}(D)$ 
in automatic differentiation frameworks such as PyTorch and JAX.

\subsection{Controlling the step size}
Vanilla gradient descent with a fixed step size tends to converge slowly 
for this problem. 
In particular, we observe overshooting when the iterates approach a local 
optimizer near the boundary of the feasible region.
To avoid this behavior, we control the step size by normalizing the gradient 
row-wise (line~\ref{alg-ln:pgd-norm-grad} in \alref{alg:pgd}) and applying a 
backtracking line search (line~\ref{alg-ln:pgd-bls}, pseudocode 
given in \alref{alg:armijo}). 
This yields good solutions to 
\eqref{eqn:singlelevel}--\eqref{eqn:singlelevel-inequal} in as few as
$n_\mathrm{pgd} = 10$ descent steps. 
We note that the \emph{signed} gradient has been used as alternative to 
the gradient when generating adversarial examples 
\citep{goodfellow2015explaining, madry2018towards}.
Both are reasonable approaches---which is more effective likely depends on 
the geometry of the problem \citep{balles2020geometry}.

\begin{algorithm}
    \begin{algorithmic}[1]
      \Procedure{BLS}{$\eta, \mat{X}, \Delta \mat{X}, \mat{G}$}
        \State $t \gets c \langle \mat{G}, \Delta \mat{X} \rangle_F$ \Comment{Frobenius inner prod.}
        \State $\tilde{\mat{X}} \gets \mat{X} - \eta \Delta \mat{X}$
        \While {$f(\mat{X}) - f(\tilde{\mat{X}}) < \eta t$}
          \State $\eta \gets \tau \eta$ \Comment{Reduce step size}
          \State $\tilde{\mat{X}} \gets \mat{X} - \eta \Delta \mat{X}$
        \EndWhile
        \State \textbf{return} $\eta$
      \EndProcedure
    \end{algorithmic}
    \caption{Backtracking line search for \alref{alg:pgd}. 
      Following \citet{armijo1966minimization} we set $\tau = c = \frac{1}{2}$.}
    \label{alg:armijo}
  \end{algorithm}

\subsection{Projection operator}
PGD depends crucially on an efficient implementation of the projection operator  
\begin{equation}
  \textsc{Proj}(\mat{X}_\pois) = \arg \min_{\mat{X} \in \Omega} \| \mat{X}_\pois - \mat{X} \|_F \label{eqn:proj}
\end{equation}
where $\Omega = \bigcap_{j = 1}^{J} \Omega_j$ and 
$\Omega_j = \{\mat{X}_\pois \in \reals^{m \times d}: g_j(\data_\pois) \leq 0\}$ 
is the set that satisfies the $j$-th inequality constraint. 
Since $\Omega_j$ is convex by \eqref{eqn:inequal-p-ball}, we can solve 
\eqref{eqn:proj} using Dykstra's algorithm.
It finds an asymptotic solution by recursively projecting onto each 
convex set $\Omega_j$. 
Pseudocode for Dykstra's algorithm is provided in \alref{alg:proj}, 
based on the stopping condition by \citet{birgin2005robust}. 
Note that it calls the projection operator $\Pi_j$ onto $\Omega_j$, which 
can be implemented efficiently for the norm-ball constraints in 
\eqref{eqn:inequal-p-ball}.
For instance, \citet{duchi2008efficient} provides a linear-time projection 
algorithm onto the $\ell_1$-ball.

\begin{algorithm}
  \begin{algorithmic}[1]
    \Procedure{Proj}{$\mat{X}$}
      \State $\mat{Y}_j \gets \mathbf{0}^\top$ \Comment{Initialize increments}
      \State $c_I = \infty$
      \While {$c_I \geq \psi \wedge i < n_\mathrm{proj}$}
        \State $c_I \gets 0; \ i \gets i + 1$
        \For {$j \in \{1, \ldots, J\}$}
          \State $\tilde{\mat{Y}} \gets \mat{Y}_j$
          \State $\mat{Z} \gets \mat{X} - \tilde{\mat{Y}}$
          \State $\mat{X} \gets \Pi_j(\mat{Z})$ \Comment{Update iterate} \label{alg-ln:proj-j}
          \State $\mat{Y}_j \gets \mat{X} - \mat{Z}$ \Comment{Update increment}
          \State $c_I \gets c_I + \|\tilde{\mat{Y}} - \mat{Y}_j \|_F^2$ \Comment{Update stopping cond.}
        \EndFor
      \EndWhile
      \State \textbf{return} $\mat{X}$
    \EndProcedure
  \end{algorithmic}
  \caption{Projection operator onto the intersection of $J$ convex sets 
    $\Omega = \bigcap_{j = 1}^{J} \Omega_j$
    based on Dykstra's algorithm \citep{birgin2005robust}. 
    Here $\psi$ is the tolerance for the stopping condition, 
    $n_\mathrm{proj}$ is the maximum number of iterations, 
    and $\Pi_j$ is the projection operator onto $\Omega_j$.}
  \label{alg:proj}
\end{algorithm}

\section{Experimental details}
\label{app:expt-detail}

In this appendix, we provide further details about the experiments presented 
in \secref{sec:experiments}.

\subsection{Implementation and hardware}
We implemented our experiments in JAX version 0.2.5 \citep{jax2018github} 
owing to its advanced automatic differentiation features. 
Experiments were run on a HPC cluster using a NVIDIA P100 GPU with 12GB 
GPU memory and 32GB CPU memory.
We repeated all simulations 100~times and reported mean values to 
account for multiple sources of randomness.

\subsection{Learning algorithm} 
We assumed the organization and the attacker both used L-BFGS 
\citep{liu1989limited} to fit their models. 
We stopped L-BFGS after a maximum of 1000 iterations or when the 
$\ell_\infty$-norm of the gradient reached a tolerance of $10^{-6}$---whichever 
occurred first.
For the multi-class case, we used a one-vs-rest strategy, since the learning 
problem \eqref{eqn:cr-risk} does not allow for vector-valued labels. 
Inputs to the model were normalized by their $\ell_2$-norm to ensure the 
condition for \eqref{eqn:grnb} was satisfied.

\subsection{Attacker settings}
We configured the attacker to run PGD (see Appendix~\ref{app:pgd}) for 
$n_\mathrm{pdg} = 10$ iterations using an initial step size $\eta_0 = d/10$. 
The projection operator was run with a tolerance of $\psi = 10^{-6}$ 
for a maximum of $n_\mathrm{proj} = 50$ iterations. 

The fraction of poisoned instances in the training set $|\data_\pois|/|\data|$ 
varied across the experiments (see Table~\ref{tbl:num-pois}), but was always 
less than 1\%.
For a given experiment, the number of poisoned instances $|\data_\pois|$ was 
set to be large enough so that only poisoned instances were unlearned. 

Unless otherwise specified, we used the influence norm cost function 
\eqref{eqn:cost-infl} and ignored model dependence on the poisoned data. 
We found these settings gave a reasonable trade-off between attack time and 
poison quality.

\begin{table*}
  \centering
  \caption{Number of poisoned instances $|\data_\pois|$ for experiments in Sec.~\ref{sec:expt-results}.}
  \label{tbl:num-pois}
  \ifarxiv \else \footnotesize \fi
  \begin{tabular}{lrrrrrr}
  \toprule
  Dataset       & $\lambda$ & $\sigma$ &      $p$ &     $r$ & Cost function & $|D_\mathrm{psn}|$ \\
  \midrule
  Fashion-MNIST & $10^{-3}$ &       10 &        1 &  $d/20$ &     Influence &                 50 \\
  HAR           & $10^{-3}$ &       10 &        1 &  $d/20$ &     Influence &                 20 \\
  MNIST         & $10^{-3}$ &       10 &        1 &  $d/20$ &     Influence &                 40 \\
  Binary-MNIST  & $10^{-2}$ &       10 &        1 &  $d/20$ &     Influence &                600 \\
  Binary-MNIST  & $10^{-3}$ &       10 &        1 &  $d/20$ &     Influence &                 20 \\
  Binary-MNIST  & $10^{-3}$ &       10 &        1 &  $d/20$ &      Gradient &                 20 \\
  Binary-MNIST  & $10^{-3}$ &       10 &        1 &  $d/20$ &          GRNB &                 20 \\
  Binary-MNIST  & $10^{-3}$ &       10 &        1 & $d/200$ &     Influence &                 90 \\
  Binary-MNIST  & $10^{-3}$ &       10 &        1 &   $d/2$ &     Influence &                 20 \\
  Binary-MNIST  & $10^{-3}$ &       10 &        1 &     $d$ &     Influence &                 10 \\
  Binary-MNIST  & $10^{-3}$ &       10 & $\infty$ &  $1/20$ &     Influence &                100 \\
  Binary-MNIST  & $10^{-3}$ &       10 & $\infty$ &  $1/10$ &     Influence &                 60 \\
  Binary-MNIST  & $10^{-3}$ &       10 & $\infty$ &   $1/2$ &     Influence &                 10 \\
  Binary-MNIST  & $10^{-3}$ &       10 & $\infty$ &       1 &     Influence &                 10 \\
  Binary-MNIST  & $10^{-4}$ &       10 &        1 &  $d/20$ &     Influence &                  5 \\
  Binary-MNIST  & $10^{-5}$ &       10 &        1 &  $d/20$ &     Influence &                  1 \\
  Binary-MNIST  & $10^{-2}$ &        1 &        1 &  $d/20$ &     Influence &                 60 \\
  Binary-MNIST  & $10^{-3}$ &        1 &        1 &  $d/20$ &     Influence &                  5 \\
  Binary-MNIST  & $10^{-4}$ &        1 &        1 &  $d/20$ &     Influence &                  1 \\
  Binary-MNIST  & $10^{-5}$ &        1 &        1 &  $d/20$ &     Influence &                  1 \\
  \bottomrule
  \end{tabular}
\end{table*}

We applied $J = 2$ inequality constraints of the form given in 
\eqref{eqn:inequal-p-ball}: 
\begin{align*}
  g_1(\data_\pois) &= \sup_{\vec{x} \in \mathrm{rows}(\mat{X}_\pois - \mat{V}_1)} 
    \left\| \vec{x} \right\|_{\infty} - r_1 &\leq 0, \\
  g_2(\data_\pois) &= \sup_{\vec{x} \in \mathrm{rows}(\mat{X}_\pois - \mat{X}_\mathrm{ref})} 
    \left\| \vec{x} \right\|_{p} - r &\leq 0.
\end{align*}
The first constraint ensures features are within valid ranges. 
For MNIST and Fashion-MNIST, we set $\mat{V}_1 = \mat{0}_{m,d}$ (zero matrix) 
and $r_1 = 1$ to ensure pixel intensities were in the range $[0, 1]$.
For HAR, we set $\mat{V}_1 = -\mat{1}_{m,d}$ 
(all-ones matrix) and $r_1 = 2$ to ensure features were in the range $[-1, 1]$.
The second constraint bounds the attacker's perturbations to the clean 
reference features $\mat{X}_\mathrm{ref}$. 
Unless otherwise specified we set $p = 1$ and $r = d / 20$.

\subsection{Organization settings}
We set $\epsilon = 1$ and $\delta = 10^{-4}$ to achieve a reasonable 
erasure guarantee.
By default, we applied moderate regularization with $\lambda = 10^{-3}$ 
and perturbed the objective with noise of magnitude $\sigma = 10$. 
These settings were found to balance model accuracy and unlearning 
efficiency in a benign setting (see Table~\ref{tbl:retrain-int-sigma-lamb}).

\section{Additional results on long-term effectiveness}

This appendix includes additional experimental results on the long-term 
effectiveness of our attack. 
In the previous results (Sec.~\ref{sec:expt-results} excluding 
Figure~\ref{fig:mnist-binary-prolonged-attack}), we measured the effectiveness 
of our attack in terms of the \emph{retrain interval}---the number of erasure 
requests handled before full retraining is triggered for the first time. 
We now consider the impact of our attack when erasure requests continue to be 
processed, even after retraining occurs potentially many times.
For these experiments we report the \emph{retrain frequency}---the fraction 
of erasure requests that trigger retraining---as a measure of effectiveness. 
A higher retrain frequency means our attack is more effective at increasing 
the computational cost of unlearning.

We report the retrain frequency averaged over 500 erasure requests, where 
all 500 requests are poisoned (i.e., we fix $|\data_\pois| = 500$). 
This yields precise estimates for most experiments, where the average retrain 
interval is significantly less than 500 and the retrain frequency is 
significantly larger than 1. 

Table~\ref{tbl:retrain-freq-sigma-lamb} reports long-term effectiveness 
as a function of the model sensitivity parameters, similarly to 
Table~\ref{tbl:retrain-int-sigma-lamb}.
We observe that our attack yields the largest gains in retrain frequency 
when retraining would be infrequent in a benign setting. 
In other words, our attack is most effective in the regime where unlearning 
would normally be most effective. 
We note that there is some degree of uncertainty in estimates of retrain 
frequency close to zero---i.e., for $\lambda \geq 10^{-3}$.

We examined the retrain frequency as a function of the perturbation 
constraint ($p$ and $r$), however we omit the results here as they 
qualitatively similar to Table~\ref{tbl:retrain-int-radius}.
Tables~\ref{tbl:retrain-freq-surrogate}--\ref{tbl:retrain-freq-data} 
report results on alternative cost functions, transferability and 
additional datasets.
These results are analogous to 
Tables~\ref{tbl:retrain-int-surrogate}--\ref{tbl:retrain-int-data} in the 
main paper and show similar trends in terms of attack effectiveness.

\begin{table}[t]
  \caption{Long-term attack effectiveness on Binary-MNIST as a function of the 
  regularization strength $\lambda$ and magnitude of the objective 
  perturbation $\sigma$.
  The accuracy is reported for the initial model, prior to processing 
  erasure requests.}
  \label{tbl:retrain-freq-sigma-lamb}
  \centering
  \ifarxiv \else \footnotesize \fi 
  \begin{tabularx}{\ifarxiv 0.75\linewidth \else \linewidth \fi}{ccRRRRr}
    \toprule
    $\sigma$ & $\lambda$ & \multicolumn{2}{c}{Accuracy} & \multicolumn{3}{c}{Retrain frequency (\%)} \\
    \cmidrule(lr){3-4} \cmidrule(lr){5-7}
             &           &       Benign &        Attack &      Benign &      Attack &  \% $\uparrow$ \\
    \midrule
    \multirow{4}{*}{1} 
             & $10^{-5}$ &        0.962 &         0.962 &        21.7 &        27.4 &     26.1 \\
             & $10^{-4}$ &        0.968 &         0.968 &        16.3 &        37.8 &      132 \\
             & $10^{-3}$ &        0.958 &         0.952 &        5.65 &        53.6 &      849 \\
             & $10^{-2}$ &        0.927 &         0.921 &       0.404 &        1.88 &      366 \\
    \midrule
    \multirow{4}{*}{10}
             & $10^{-5}$ &        0.919 &         0.917 &         100 &         100 &    -0.01 \\
             & $10^{-4}$ &        0.933 &         0.928 &        9.67 &        26.8 &      177 \\
             & $10^{-3}$ &        0.955 &         0.947 &       0.656 &        7.69 &     1070 \\
             & $10^{-2}$ &        0.926 &         0.920 &           0 &       0.132 &       -- \\
    \bottomrule
  \end{tabularx}
\end{table}

\begin{table}[t]
  \caption{Long-term attack effectiveness and computation time for different 
    choices of the cost function.}
  \label{tbl:retrain-freq-surrogate}
  \centering
  \ifarxiv \else \footnotesize \fi
  \begin{tabularx}{\ifarxiv 0.75\linewidth \else \linewidth \fi}{lLRR}
    \toprule
    Cost function & Ignore model dep. & Retrain frequency & Attack time (s) \\
    \midrule
    \multirow{2}{*}{GRNB \eqref{eqn:cost-grnb}}      
              &                      No &            7.91 &           170.5 \\
              &                     Yes &            8.05 &            22.6 \\
    \midrule
    \multirow{2}{*}{Influence norm \eqref{eqn:cost-infl}} 
              &                      No &            5.07 &            62.4 \\
              &                     Yes &            7.69 &            8.84 \\
    \midrule
    \multirow{2}{*}{Gradient norm \eqref{eqn:cost-grad}}
              &                      No &            4.74 &            21.8 \\
              &                     Yes &            4.64 &            3.87 \\
    \bottomrule 
  \end{tabularx}
\end{table}

\begin{table}[t]
  \caption{Long-term effectiveness of the attack in white-box versus grey-box settings.}
  \label{tbl:retrain-freq-transfer}
  \centering
  \ifarxiv \else \footnotesize \fi
  \begin{tabularx}{\ifarxiv 0.75\linewidth \else \linewidth \fi}{lRR}
    \toprule
    Dataset      & \multicolumn{2}{c}{Retrain frequency (\%)} \\
    \cmidrule(lr){2-3}
                 & Surrogate data (Grey-box) & Training data (White-box) \\
    \midrule
    MNIST        &                      4.63 &                      4.43 \\
    Binary-MNIST &                      7.90 &                      7.69 \\
    \bottomrule
  \end{tabularx}
\end{table}

\begin{table}[t]
  \caption{Long-term effectiveness of the attack for various datasets.}
  \label{tbl:retrain-freq-data}
  \centering
  \ifarxiv \else \footnotesize \fi
  \begin{tabularx}{\ifarxiv 0.75\linewidth \else \linewidth \fi}{lRRRRr}
    \toprule
    Dataset        & \multicolumn{2}{c}{Initial accuracy} & \multicolumn{3}{c}{Retrain frequency (\%)} \\
    \cmidrule(lr){2-3} \cmidrule(lr){4-6}
                   &       Benign &        Attack &      Benign &      Attack &  \% $\uparrow$ \\
    \midrule    
    MNIST          &        0.867 &         0.867 &        1.27 &        4.43 &     250 \\
    Binary-MNIST   &        0.955 &         0.947 &       0.656 &        7.69 &    1070 \\
    Fashion-MNIST  &        0.756 &         0.755 &       0.778 &        2.18 &     180 \\
    HAR            &        0.837 &         0.836 &        2.37 &        9.37 &     296 \\
    \bottomrule
  \end{tabularx}
\end{table}

\end{document}